\documentclass[letterpaper, 10 pt, journal, twoside]{ieeetran}
\usepackage{amsmath,amsfonts}
\usepackage{algorithmic}
\usepackage{algorithm}
\usepackage{array}
\usepackage[caption=false,font=normalsize,labelfont=sf,textfont=sf]{subfig}
\usepackage{textcomp}
\usepackage{stfloats}
\usepackage{url}
\usepackage{verbatim}
\usepackage{graphicx}
\usepackage{cite}
\usepackage{threeparttable} 
\usepackage{booktabs}  
\usepackage{tikz} 
\usepackage{xcolor} 
\usepackage{hyperref} 

\newcommand{\startsymbol}{\protect\tikz[baseline=-0.6ex]\protect\draw[fill=orange] (0,0) circle (0.12cm); }
\newcommand{\coarsesymbol}{\protect\tikz[baseline=-0.6ex]\protect\draw[fill=magenta!70] (0,0) circle (0.12cm); }
\newcommand{\optsymbol}{\protect\tikz[baseline=-0.6ex]\protect\draw[fill=blue!60] (0,0) circle (0.12cm); }

\begin{document}

\title{LAMP: Implicit Language Map for Robot Navigation}

\author{Sibaek Lee$^{1, 2, \dagger}$, Hyeonwoo Yu$^1$, Giseop Kim$^3$, and Sunwook Choi$^{2, *}$%

\thanks{$^{1}$Sibaek Lee and Hyeonwoo Yu are with the Department of Intelligent Robotics, Sungkyunkwan University, Suwon, South Korea. {\tt\small \{lmjlss, hwyu\}@skku.edu}}%
\thanks{$^{2}$Sibaek Lee and Sunwook Choi are with NAVER LABS, Seongnam, South Korea. {\tt\small \{sibaek.lee, sunwook.choi\}@naverlabs.com}}
\thanks{$^{3}$Giseop Kim is with the Department of Robotics and Mechatronics Engineering, DGIST, Daegu, South Korea. {\tt\small gsk@dgist.ac.kr}}%
\thanks{$^{\dagger}$Work done during an internship at NAVER LABS. $^{*}$Corresponding author: Sunwook Choi. Project page: \href{https://lab-of-ai-and-robotics.github.io/LAMP/}{https://lab-of-ai-and-robotics.github.io/LAMP/}}%
\thanks{Digital Object Identifier (DOI): 10.1109/LRA.2025.3619820}
}

\markboth{IEEE Robotics and Automation Letters. Preprint Version. Accepted October, 2025}
{Lee \MakeLowercase{\textit{et al.}}: LAMP: Implicit Language Map for Robot Navigation}

\maketitle

\begin{abstract}
Recent advances in vision-language models have made zero-shot navigation feasible, enabling robots to interpret and follow natural language instructions without requiring labeling. However, existing methods that explicitly store language vectors in grid or node-based maps struggle to scale to large environments due to excessive memory requirements and limited resolution for fine-grained planning. We introduce LAMP (Language Map), a novel neural language field-based navigation framework that learns a continuous, language-driven map and directly leverages it for fine-grained path generation. Unlike prior approaches, our method encodes language features as an implicit neural field rather than storing them explicitly at every location. By combining this implicit representation with a sparse graph, LAMP supports efficient coarse path planning and then performs gradient-based optimization in the learned field to refine poses near the goal. Our two-stage pipeline of coarse graph search followed by language-driven, gradient-guided optimization is the first application of an implicit language map for precise path generation. This refinement is particularly effective at selecting goal regions not directly observed by leveraging semantic similarities in the learned feature space. To further enhance robustness, we adopt a Bayesian framework that models embedding uncertainty via the von Mises–Fisher distribution, thereby improving generalization to unobserved regions. To scale to large environments, LAMP employs a graph sampling strategy that prioritizes spatial coverage and embedding confidence, retaining only the most informative nodes and substantially reducing computational overhead. Our experimental results, both in NVIDIA Isaac Sim and on a real multi-floor building, demonstrate that LAMP outperforms existing explicit methods in both memory efficiency and fine-grained goal-reaching accuracy, opening new possibilities for scalable, language-driven robot navigation.
\end{abstract}


\section{Introduction}
\IEEEPARstart{I}{n} robotics, map representation has significant value as it enables robots to perceive spatial information and effectively perform a wide range of tasks. Traditionally, 3D maps have been constructed by assigning coordinate values, RGB information, and semantic labels to point clouds or voxels \cite{hornung2013octomap, engel2014lsd, zhang2014loam, qi2017pointnet}. However, with the emergence of visual language models \cite{radford2021learning, caron2021emerging}, it has become possible to represent space through language embedding vectors, thereby capturing long-tailed objects and abstract concepts without explicit labeling. This shift transforms the environment from a purely geometric construct into a linguistically enriched space, opening the door to more intuitive and meaningful robot-environment interactions. Leveraging the strengths of these language embeddings, they are being applied in various fields such as 3D semantic segmentation \cite{kerr2023lerf, engelmann2024opennerf, qin2024langsplat}, path planning \cite{gadre2023cows, shah2023lm}, scene understanding \cite{azuma2022scanqa, cascante2022simvqa}, manipulation \cite{huang2022language}, extended reality \cite{kurai2025magicitem}, and autonomous driving \cite{jatavallabhula2023conceptfusion}.

However, current language map representations are limited to small environments and encounter significant challenges for large-scale deployment. Since robots must operate indoors and outdoors to perform tasks autonomously, they require effective space interaction in extensive areas. Existing language maps for path planning methods typically store language vectors explicitly in 2D grids \cite{huang2023visual, huang2023audio, gadre2023cows} or topological graph nodes \cite{shah2023lm, werby2024hierarchical, arul2024vlpg}. As environments scale up, this explicit storage becomes excessively memory and computationally intensive. Some approaches adopt sparse representations to alleviate these constraints, enabling coarse-level path planning but limiting fine-grained navigation. 
This limitation arises from the inherent difficulty of densely and explicitly storing information on large scales. To address this gap, we propose an implicit language map representation that continuously models language vectors from RGB-only input, facilitating memory-efficient path planning that supports not only coarse navigation but also fine-grained route guidance.

\begin{figure*}
    \centering
    \includegraphics[width=0.88\textwidth]{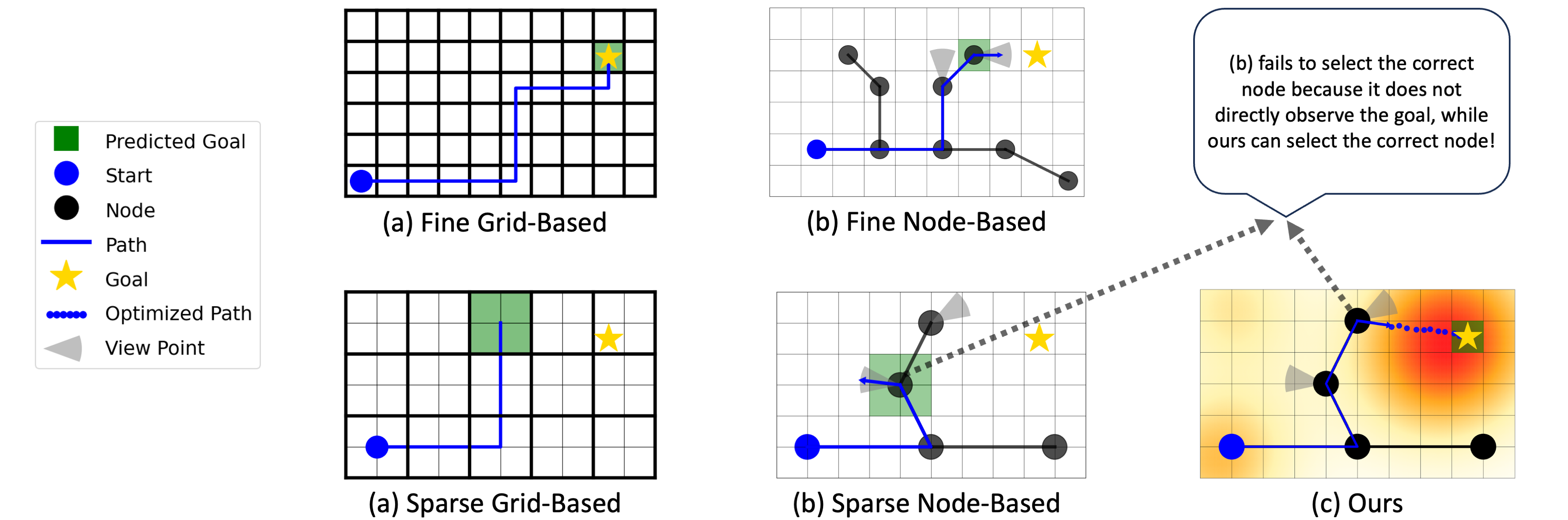}
    \caption{Comparison of three language map representation methods. (a) The grid-based approach struggles to accurately represent objects at coarse resolutions and requires excessive memory when increasing grid resolution to capture finer details. (b) The node-based approach fails to capture important object details when node spacing is too coarse and cannot guarantee precise path planning. (c) In contrast, our implicit method maintains memory efficiency even at large scales while providing fine-level path guidance.}
    \label{fig:intro}
\end{figure*}

In Fig. \ref{fig:intro}, (a) the grid-based method leverages depth images and a Visual Language Model \cite{li2022language, radford2021learning} to integrate language information into a 3D representation, which is subsequently projected onto a 2D grid map using a top-down approach \cite{huang2023visual, huang2023audio, gadre2023cows}. This method captures objects well when grid cells are smaller than the objects. However, it struggles to represent smaller objects because of the grid's coarse resolution. Moreover, a dense grid in large environments consumes too much memory, making it unsuitable for robotics.
In contrast, (b) the node-based method places nodes at key spatial positions and explicitly stores CLIP vectors and pose information \cite{shah2023lm, werby2024hierarchical, arul2024vlpg}. It uses less memory than grid-based approach and can detect smaller objects at the nodes. However, in large scenes, it is impractical to place enough nodes to capture the fine-grained details required for precise path planning. For instance, Fig. \ref{fig:intro} shows that with sparse nodes, the node-based method fails to locate the goal. Conversely, with dense nodes, the method reaches near the goal but struggles to approach the goal with fine precision.
Finally, (c) our implicit method addresses these issues by modeling language information as a continuous function using a neural network. This network infers language vectors in regions not explicitly observed during training, enabling flexible and detailed representations that continuously capture smaller objects. Consequently, the system performs precise path planning without storing vectors at every node, significantly reducing memory requirements. Furthermore, even if a node doesn't directly observe the goal, our method selects nodes near the goal, ensuring guidance to unobserved targets. By maintaining only a sparse set of nodes with pose information and generating language features on demand, our approach achieves memory efficiency and detailed mapping in large environments.

Building on the strengths of our implicit language map, we propose methods to construct and utilize this representation more effectively. Although the implicit map provides a continuous function for language vectors in unobserved areas, mapping camera poses to language vectors in a highly nonlinear manner poses a challenge. Moreover, CLIP vectors often capture multiple objects simultaneously, injecting noise into the language embeddings themselves.
To address these issues, we apply a Bayesian approach to our neural network to account for aleatoric uncertainty \cite{kendall2017uncertainties, lee2025bayesian}. By incorporating this uncertainty into our loss function, we enhance the implicit map's generalization to untrained poses. We also introduce a graph sampling strategy for large-scale graphs that utilizes both the uncertainty information derived from our Bayesian model and the network gradients. As the number of nodes increases in large-scale environments, exhaustive exploration becomes impractical, and node selection critically affects task performance \cite{zhang2023large, leskovec2006sampling}. Consequently, graph sampling is essential for efficient large-scale planning and ensures our implicit map remains compact yet effective. By modeling embedding uncertainty via a von Mises–Fisher Bayesian formulation and incorporating semantic sensitivity into node sampling, our system achieves far greater robustness than naive gradient-based optimization alone.
We summarize our main contributions of LAMP (Language Map) as follows:
 \begin{itemize}  
     \item We introduce LAMP, the first implicit language map leveraging a language-driven continuous field for fine-grained path generation using only RGB images.
     \item We incorporate a Bayesian approach to improve generalization performance to untrained poses.
     \item We develop a graph sampling technique that optimizes node selection using language features and uncertainty predictions.
 \end{itemize}

\section{Related work}
Traditional Mapping Approaches. Robots have long relied on geometric maps such as occupancy grids \cite{elfes1989using} and landmark-based \cite{davison2003real, montemerlo2002fastslam, mur2015orb} representations for navigation. While effective for path planning, these low-level depictions carry no inherent semantic information and the sparse features in SLAM maps are difficult to interpret in human terms without additional context \cite{davison2003real, montemerlo2002fastslam, mur2015orb, cadena2016past}. To make maps more human-meaningful, researchers introduced label-based semantic mapping by annotating regions with object and place names \cite{mccormac2017semanticfusion, dube2018segmap}, but this approach is confined to pre-defined classes as it depends on explicit annotations or trained detectors. Consequently, conventional maps cannot handle references to novel, long-tail concepts and thus fail to generalize to many unseen objects or descriptors.

Language-Based Mapping Approaches. To overcome these limitations, recent work fuses natural language understanding with mapping. Vision–Language Model (VLM) \cite{radford2021learning, caron2021emerging} has given rise to approaches where robots use natural language instructions to reach semantically specified targets. For instance, NeRF \cite{kerr2023lerf, engelmann2024opennerf} and Gaussian Splatting \cite{qin2024langsplat, peng20243d} integrate language embeddings into scene representations for view synthesis and semantic segmentation, but these approaches are not primarily designed for navigation and cannot scale to large environments due to high memory and rendering costs. Other navigation-oriented methods attempt to address this by explicitly storing language vectors in 2D grids \cite{huang2023visual, huang2023audio, gadre2023cows} or in topological graph nodes \cite{shah2023lm, werby2024hierarchical, arul2024vlpg}, but they still suffer from excessive memory demands and coarse planning resolution in large scale environments. To address this limitation, we introduce the first approach that implicitly represents space for memory efficiency and leverages this implicit map in an optimization framework to generate fine-grained navigation paths.

\section{Method}
We introduce a map representation that continuously encodes language features within a large-scale space, ensuring memory efficiency and enabling fine-grained path planning. The section is organised into the following parts: Section \ref{subsec:A} states the navigation problem. Section \ref{subsec:B} describes how we construct the implicit language map, incorporate uncertainty, and prune the topological graph. Section \ref{subsec:C} explains how the resulting map is used to generate coarse paths and refine them into precise goal poses.

\begin{figure*}
    \centering
    \includegraphics[width=0.96\textwidth]{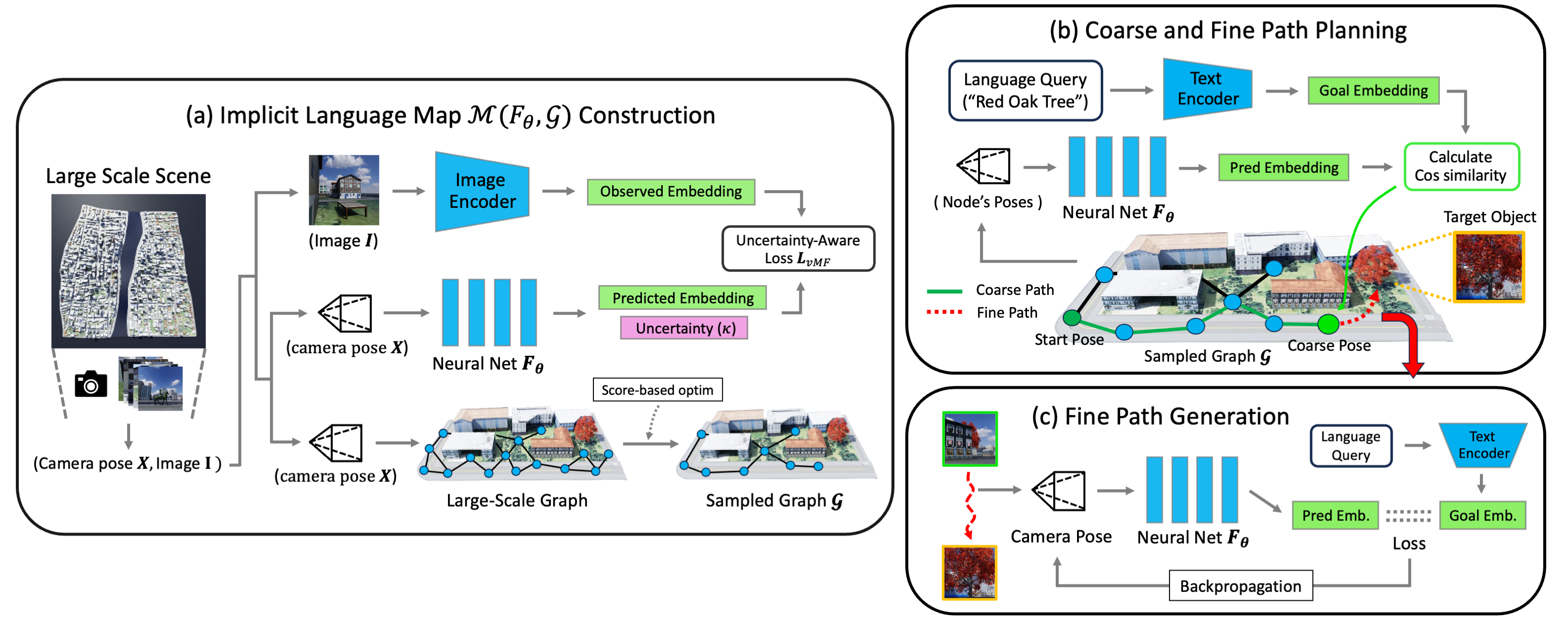}
    \caption{System Overview. (a) Implicit Language Map Construction: The robot traverses the environment and collects pairs of camera poses $\mathbf{x}$ and corresponding images $\mathbf{I}$. Neural network \(F_\Theta\) maps each pose \(\mathbf{x}\) to a language embedding \(\mathbf{z} = F_\Theta(\mathbf{x})\). Since processing the full large-scale topological graph is computationally expensive, we sample the graph \(\mathcal{G}\) using our proposed score-based optimization for coarse planning. (b) Coarse Path Planning: Given a user’s natural language query such as “red oak tree”, we encode a goal embedding and apply A* on the sampled graph $\mathcal{G}$ to obtain a coarse path to the node whose embedding best matches the goal embedding. (c) Fine Path Generation: We then generate the pose using \(F_\Theta\) to maximize cosine similarity, moving from the coarse pose to a fine pose that offers a clear view of the target object.}
    \label{fig:method}
\end{figure*}

\subsection{Problem Definition}
\label{subsec:A}
We formalize the navigation task as follows. Let \(\mathcal{M} = \{F_\Theta, \mathcal{G}\}\) denote an implicit language map of the environment, where \(F_\Theta\) is a neural network that continuously represents language features and \(\mathcal{G} = (\mathcal{V}, \mathcal{E})\) is a topological graph providing a global structure for path planning. Navigation begins from an initial pose \(\mathbf{x}_{init}\) within this static environment (i.e., \(\mathcal{M}\) does not change) and receives a natural language query \(Q\) describing a desired destination. The objective is to generate a two‑stage path: first, a coarse path \(\gamma_c = (x_0, \ldots, x_c)\) obtained by searching over the graph’s node set \(\mathcal{V}\) to reach an intermediate goal \(x_c\); second, a fine path \(\gamma_f = (x_c, \ldots, x_G)\) produced by optimizing \(F_\Theta\) to precisely reach the final goal \(x_G\). In summary, given \((\mathcal{M}, \mathbf{x}_{init}, Q)\), the system generates a coarse path \(\gamma_c\) via \(\mathcal{G}\) and then refines it into a fine path \(\gamma_f\) via \(F_\Theta\) to obtain an accurate path to the final goal \(\mathbf{x}_G\).

\subsection{Implicit language map Pipeline}
\label{subsec:B}

\subsubsection{Implicit Language Map Construction}
\label{subsubsec:B_1}
To construct \(\mathcal{M}\), we assume that the robot has traversed the environment extensively, collecting \((\mathbf{x}, \mathbf{I})\) pairs, where \(\mathbf{x}\) represents the camera pose and \(\mathbf{I}\) is the corresponding image. Inspired by NeRF frameworks~\cite{mildenhall2021nerf}, we implicitly encode the language embeddings of the scene as viewed from each position, rather than encoding the color and density at the positions themselves. Concretely, let \(\mathbf{t} \in \mathbb{R}^{3}\) be the 3D coordinate and \(\mathbf{q} \in \mathbb{R}^{4}\) be a quaternion representing orientation. We concatenate these into \(\mathbf{x} = [\mathbf{t},\, \mathbf{q}] \in \mathbb{R}^{7}.\) Our neural network \(F_\Theta\) then maps \(\mathbf{x}\) to a \(d\)-dimensional CLIP embedding:
\[
    F_\Theta(\mathbf{x}) = \mathbf{z} \in \mathbb{R}^{d},
\]
where \(\mathbf{z}\) captures the language features observed in the input image \(\mathbf{I}\). By training on \((\mathbf{x}, \mathbf{I})\) pairs, \(F_\Theta\) learns a continuous language representation of the environment using only RGB images.

However, relying solely on \(F_\Theta\) for path planning can lead to local minima issues. To address this, we use the topological graph \(\mathcal{G}\), whose nodes \(v \in \mathcal{V}\) store only pose information \(\mathbf{x}_v\), from which language embeddings can be generated on demand via \(F_\Theta(\mathbf{x}_v)\). This approach contrasts with previous methods that explicitly store precomputed embeddings at every node, leading to large memory usage. By dynamically generating embeddings through \(F_\Theta\), our method significantly reduces storage while preserving language features.

\subsubsection{Bayesian Approach for Robust Embedding Predictions}
\label{subsubsec:B_2}
Although large-scale data collection allows the robot to learn from a wide set of \((\mathbf{x}, \mathbf{I})\) pairs, predicting embedding vectors for unseen poses remains challenging. Moreover, CLIP-based embeddings represent the collective semantics of multiple objects within an image, which can introduce noise when focusing on a specific object. Consequently, accounting for uncertainty is necessary for robust predictions \cite{kendall2017uncertainties, gal2016dropout}. Since each embedding vector lies on the unit hypersphere in \(\mathbb{R}^d\), the von Mises-Fisher (vMF) distribution naturally captures its directional uncertainty while preserving the unit norm. We enforce this unit-length condition by $\ell_2$-normalising every CLIP feature and network output, so cosine similarity is a true metric and the embeddings reside on the domain of the vMF distribution. To address this issue, we adopt a Bayesian approach based on the vMF distribution \cite{hasnat2017mises, nakadai2020sound}, wherein each predicted embedding \(\mathbf{z}\) lies on the unit hypersphere in \(\mathbb{R}^d\) with density:
\begin{equation}
    p(\mathbf{z} \mid \boldsymbol{\mu}, \kappa)
    \;=\;
    C_d(\kappa) \,\exp\bigl(\kappa\, \boldsymbol{\mu}^\top \mathbf{z}\bigr),
    \label{eq:vmf_likelihood}
\end{equation}
where \(\boldsymbol{\mu}\in\mathbb{R}^d\) is also a unit vector, \(\kappa \ge 0\) is the concentration parameter, and \(C_d(\kappa)\) is the normalization constant.\footnote{%
  \(C_d(\kappa) = \frac{\kappa^{d/2 - 1}}{(2\pi)^{d/2} \, I_{d/2 - 1}(\kappa)}\),
  where \(I_{\nu}\) is the modified Bessel function of the first kind.
} Intuitively, \(\kappa\) governs how tightly \(\mathbf{z}\) concentrates around the mean direction \(\boldsymbol{\mu}\). To prevent \(\kappa\) from diverging, we place a Gamma prior on it:
\begin{equation}
    p(\kappa)
    \;=\;
    \frac{\beta^\alpha}{\Gamma(\alpha)}
    \,\kappa^{\alpha-1}\,
    e^{-\beta \kappa},
    \label{eq:gamma_prior}
\end{equation}

where \(\alpha\) and \(\beta\) are hyperparameters, and \(\Gamma(\alpha)\) is the gamma function. This prior enforces reasonable bounds on the concentration parameter, mitigating overconfidence in the embedding’s direction. To implement this, we define our network $F_\Theta$ to output the two parameters of the vMF distribution for a given pose $\mathbf{x}$: the mean direction $\boldsymbol{\mu}$ and the concentration $\kappa$. Let us denote the network outputs as $(\boldsymbol{\mu}_\Theta(\mathbf{x}), \kappa_\Theta(\mathbf{x})) = F_\Theta(\mathbf{x})$. By combining the vMF likelihood in Eq.~\eqref{eq:vmf_likelihood} with the Gamma prior in Eq.~\eqref{eq:gamma_prior}, the posterior over the network parameters $\theta$ is proportional to:
\begin{equation}
    p(\theta \mid \mathbf{x}, \mathbf{z}_{\text{obs}})
    \;\propto\;
    p(\mathbf{z}_{\text{obs}} \mid F_\Theta(\mathbf{x}))\,
    p(\kappa_\Theta(\mathbf{x})),
    \label{eq:posterior}
\end{equation}
and we train the network by minimizing the negative log-posterior, which serves as our loss function for a given training pair $(\mathbf{x}, \mathbf{z}_{\text{obs}})$:
\begin{equation}
    \mathcal{L}_{\text{vMF}}(\theta; \mathbf{x}, \mathbf{z}_{\text{obs}})
    \;=\;
    -\ln p(\mathbf{z}_{\text{obs}} \mid F_\Theta(\mathbf{x}))
    \;-\;
    \ln p(\kappa_\Theta(\mathbf{x})).
    \label{eq:vmf_loss}
\end{equation}
This loss function trains the network to produce parameters $(\boldsymbol{\mu}_\Theta(\mathbf{x}), \kappa_\Theta(\mathbf{x}))$ that balance closeness to the observed embedding $\mathbf{z}_{\text{obs}}$ with regularization from the Gamma prior, thereby reducing the influence of noisy data and enhancing robustness in unobserved scenarios. By introducing a Bayesian framework on the outputs of \(F_\Theta\), our method becomes robust to noisy embeddings and provides reliable predictions in unobserved regions. This enhances path‐planning performance and also feeds into the subsequent graph‐optimization step.

\subsubsection{Node Optimization in Large-Scale Graph}
\label{subsubsec:B_3}
While our topological graph \(\mathcal{G}\) encodes the global structure for path planning, a large number of nodes can lead to computational inefficiencies during graph search \cite{zhang2023large, leskovec2006sampling}. To address this, we propose a graph sampling method that retains only the most informative nodes, scored by three criteria. Formally, for each node \(v\in\mathcal{V}\), we define:
\begin{equation}
    \mathrm{score}(v) = w_{1}\,s_{\text{vc}}(v) + w_{2}\,s_{u}(v) + w_{3}\,s_{\text{ss}}(v),
    \label{eq:score_combined}
\end{equation}
where \(s_{\text{vc}}(v)\) is a \emph{View Coverage Score}, \(s_{u}(v)\) is an \emph{Uncertainty Score}, and \(s_{\text{ss}}(v)\) is a \emph{Semantic Sensitivity Score}; the weights \(w_{1}\), \(w_{2}\), and \(w_{3}\) are hyperparameters that balance the contribution of each score to the overall score.

\begin{itemize}
    \item View Coverage Score \(s_{\text{vc}}(v)\):
    For each node \(v\), we compute the dot product between its quaternion and those of nodes in its two‑hop neighborhood to measure orientation difference. A lower similarity indicates a novel viewpoint, which enhances coverage in the sampled graph.

    \item Uncertainty Score \(s_{u}(v)\):
    Leverages our Bayesian approach to quantify aleatoric uncertainty. The score is defined as the concentration parameter, $\kappa$, of the predicted von Mises-Fisher distribution, where a higher $\kappa$ indicates greater confidence in the embedding.

    \item Semantic Sensitivity Score \(s_{\text{ss}}(v)\):
    Measures the gradient of \(F_\Theta\) with respect to the node’s pose $\mathbf{x}_v$ by computing its norm. A larger gradient suggests that small changes in pose result in substantial variation in the predicted embedding, implying a high semantic contrast. These nodes are valuable for capturing critical transition regions.
\end{itemize}

Nodes are ranked by their combined score \eqref{eq:score_combined}, and only the top-scoring nodes are retained in the sampled graph. This selective pruning reduces the computational cost of graph-based planning while preserving coverage of key viewpoints, embedding reliability, and semantic diversity. As a result, our method scales efficiently to complex environments without compromising navigation and exploration capabilities.

\subsection{Coarse to Fine Path Planning}
\label{subsec:C}

\subsubsection{Coarse Path Planning}
\label{subsec:C_1}
Given a natural language query, we first convert it into a goal embedding \(\mathbf{z}_{\text{goal}} = \mathrm{CLIP}(Q)\). For each node \(v \in \mathcal{V}\) with a stored pose \(\mathbf{x}_v\), we obtain an embedding \(F_\Theta(\mathbf{x}_v)\) and compute its cosine similarity to \(\mathbf{z}_{\text{goal}}\):
\[
    \mathrm{sim}\!\bigl(\mathbf{z}_{\text{goal}},\, F_\Theta(\mathbf{x}_v)\bigr), 
    \quad \forall v \in \mathcal{V}.
\]
The node with the highest similarity serves as our coarse semantic goal, and its pose is set to \(\mathbf{x}_c\) for the fine‐level optimization. We then employ an A* algorithm to plan a path from the robot’s current node to this coarse goal node, achieving coarse-level navigation to the language goal.

\subsubsection{Fine Path Generation}
\label{subsec:C_2}
Although \(\mathcal{G}\) provides a coarse path from the most similar node in large-scale environments, additional refinement is necessary to achieve fine-grained goals beyond the graph's node resolution. Therefore, we initiate optimization from the coarse pose \(\mathbf{x}_c\). To reduce the risk of converging to a poor local optimum, we sample multiple candidate poses near \(\mathbf{x}_c\) and select the pose with the highest similarity to \(\mathbf{z}_{\text{goal}}\), denoted as \(\mathbf{x}_{\mathrm{best}}\), which guides the subsequent optimization. We then define a small correction term \(\delta \mathbf{x}\) on top of \(\mathbf{x}_c\) and formulate the following objective:
\begin{equation}
\max_{\delta \mathbf{x}}
\Bigl[
  \mathrm{sim}\bigl(F_\Theta(\mathbf{x}_c + \delta \mathbf{x}),\, \mathbf{z}_{\text{goal}}\bigr)
  \;-\;
  \lambda_{\mathrm{dist}}\,\mathrm{dist}\bigl(\mathbf{x}_{\mathrm{best}},\, \mathbf{x}_c + \delta \mathbf{x}\bigr)
\Bigr].
\label{eq:fg-objective}
\end{equation}

Here $\mathrm{dist}(\mathbf{x}_1,\mathbf{x}_2)$ denotes the Euclidean distance in~$\mathbb{R}^3$.
We solve \eqref{eq:fg-objective} via a gradient-based method, updating \(\delta \mathbf{x}\) at each step \(t\) by
\begin{equation}
    \delta \mathbf{x}_{t+1}
    \;=\;
    \delta \mathbf{x}_t
    \;-\;
    \alpha \,\frac{\hat{m}_t}{\sqrt{\hat{v}_t} + \epsilon},
\end{equation}
where \(\hat{m}_t\) and \(\hat{v}_t\) are Adam’s bias-corrected first and second moment estimates, \(\alpha\) is the learning rate, and \(\epsilon\) is a small constant.
This process yields a refined pose \(\mathbf{x}_c + \delta \mathbf{x}^*\) that simultaneously maximizes semantic similarity and remains close to the best sampled pose \(\mathbf{x}_{\mathrm{best}}\), thus mitigating local minima. By combining coarse-level graph navigation with this guided optimization in \(F_\Theta\), we achieve fine-grained language guidance without sacrificing large-scale planning and memory efficiency.

\section{Experiments}
The purpose of our experiments is to demonstrate that LAMP, our method which implicitly incorporates language information within large-scale scenes, achieves memory efficiency and enables fine-grained zero-shot navigation. In the following subsections, Section \ref{subsec_ex:A} describes the dataset configuration and implementation details, Section \ref{subsec_ex:B} presents the experimental results obtained in the Nvidia Isaac simulation environment along with a discussion, and Section \ref{subsec_ex:C} validates our approach on a real robot by demonstrating navigation across multiple floors of a building using only a very sparse set of nodes.

\subsection{Experimental Setup}
\label{subsec_ex:A}

\subsubsection{Dataset Configuration}
\begin{figure}
    \centering
    \includegraphics[width=0.46\textwidth]{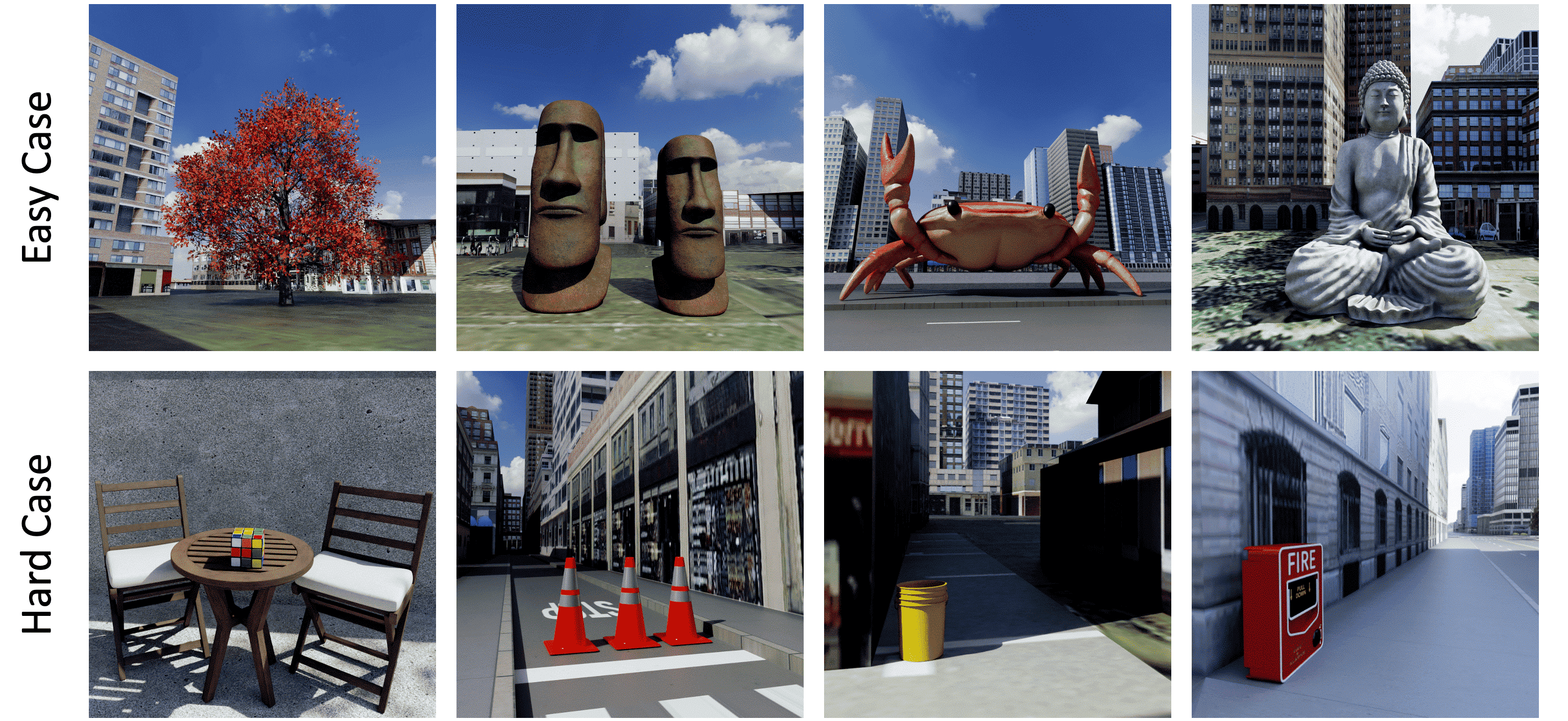}
    \caption{Examples of objects used in our simulation navigation experiments. The top row displays large objects (volume $\geq$ 1 m$^3$) such as statues and a red oak tree, while the bottom row shows smaller objects (volume $<$ 1 m$^3$) such as a Rubik’s cube or a fire alarm, which are harder to detect in a large-scale environment.}
    \label{fig:dataset}
\end{figure}

To validate our method’s memory efficiency, we evaluate our system on a large‐scale environment rather than limiting it to room‐scale datasets such as MP3D \cite{chang2017matterport3d}. Therefore, for our simulation experiments, we use the NVIDIA Isaac Sim simulator along with the City Tower Demo 3D Models Pack, which covers about 1.6~km $\times$ 1.8~km. We integrate assets from the NVIDIA library to set up a zero-shot navigation task. As shown in Fig. \ref{fig:dataset}, these assets are divided into two groups according to size: objects with a volume of 1~m$^3$ or greater are labeled as easy cases, while those with a smaller volume are hard cases. The easy cases include large items like statues and trees that are visible from many viewpoints, making them easier to navigate to. In contrast, hard cases include smaller items, such as a Rubik’s cube or a fire alarm, whose compact size makes them difficult to detect in a large-scale environment.
We assume that the robot explores the environment for an extended period to collect data. Since the data are obtained directly through the robot's own exploration, the gathered pose data ensure the robot's traversability. We collect five million pose–image pairs to provide diverse viewpoints for a refined map. 
In our real-world experiments, we deployed a robot equipped with six cameras to navigate each floor of the building. On each floor, the robot collected between 10,000 and 20,000 pose-image pairs, ensuring comprehensive coverage of the environment. The building consists of 28 floors, each spanning approximately 100~m $\times$ 70~m.


\subsubsection{Implementation Details}
In implementing our approach, we follow the MLP architecture used in NeRF~\cite{mildenhall2021nerf} by incorporating positional encoding and skip connections to capture necessary variations. Our model outputs a 512-dimensional language embedding vector along with a concentration parameter to indicate uncertainty. During training, we model uncertainty by placing a Gamma prior on the concentration parameter with $\alpha = 2$ and $\beta = 0.5$. As described in Section~\ref{subsubsec:B_3}, key nodes within the graph are selected using a heuristic weighting scheme with $w_{1} = 1$, $w_{2} = 1$, and $w_{3} = 0.5$. For the fine path generation objective (Eq. \ref{eq:fg-objective}), the distance penalty hyperparameter $\lambda_{\mathrm{dist}}$ is set to 5. We assess navigation performance using three metrics. First, \emph{success rate} is computed considering only the top $1\%$ of the predictions; a trial is deemed successful if the robot ends up within 20\,m of the center of an object. Next, the \emph{Success weighted by Path Length (SPL)} metric evaluates navigation efficiency by penalizing unnecessarily long paths. Finally, for successful trials, we calculate the \emph{goal-to-distance} measure by averaging the remaining distance from the target.

\subsection{Simulation Experiments and Analysis}
\label{subsec_ex:B}

\begin{table*}[t!]
    \centering
    \begin{threeparttable}
        \caption{Quantitative Comparison of Language Map Representations (Uniform Memory Setting)}
        \label{tab:comparison_methods1}
        \begin{tabular}{lcccccccc}
            \toprule 
            \textbf{Method}     & \textbf{Memory(GB)} & \textbf{Time (s)} & \textbf{SR (Easy)} & \textbf{SPL (Easy)} & \textbf{GDist (Easy)} & \textbf{SR (Hard)} & \textbf{SPL (Hard)} & \textbf{GDist (Hard)} \\ 
            \midrule
            Grid-based (Sparse) & 0.055               & 0.0001            & 0.08               & 0.07                & 15.50                 & 0.0                & 0.0                 & --                    \\ 
            Node-based (Sparse) & 0.057               & 0.0066            & 0.41               & 0.36                & 12.69                 & 0.21               & 0.17                & 4.14                  \\ 
            Ours (Implicit)     & 0.057               & 0.8041            & \textbf{0.67}      & \textbf{0.62}       & \textbf{6.36}         & \textbf{0.42}      & \textbf{0.38}       & \textbf{1.74}         \\ 
            \bottomrule
        \end{tabular}
    \end{threeparttable}
\end{table*}

\begin{table*}[t!]
    \centering
    \begin{threeparttable}
        \caption{Quantitative Comparison of Language Map Representations (Memory-Intensive Setting)}
        \label{tab:comparison_methods2}
        \begin{tabular}{lcccccccc}
            \toprule 
            \textbf{Method}      & \textbf{Memory(GB)} & \textbf{Time (s)} & \textbf{SR (Easy)} & \textbf{SPL (Easy)} & \textbf{GDist (Easy)} & \textbf{SR (Hard)} & \textbf{SPL (Hard)} & \textbf{GDist (Hard)} \\ 
            \midrule
            Grid-based (Dense)   & 56.34               & 0.1036            & 0.83               & 0.80                & 2.44                  & 0.36               & 0.33                & 1.27                  \\ 
            Node-based (Dense)   & 3.962               & 0.3395            & 0.67               & 0.61                & 7.51                  & 0.47               & 0.41                & 2.38                  \\ 
            Ours (Implicit)      & \textbf{0.057}      & 0.8041            & 0.67               & 0.62                & 6.36                  & 0.42               & 0.38                & 1.74                  \\ 
            \bottomrule
        \end{tabular}
        This experiment uses the NVIDIA's City Tower Demo 3D Models Pack scene, which covers 1.6\,km $\times$ 1.8\,km. In the Grid-based (Sparse) method, the grid cell is set to 12m, while the Grid-based (Dense) method utilizes a grid size of 0.4m. The Node-based (Sparse) map comprises 30,000 nodes, and the Node-based (Dense) map contains 2,000,000 nodes. Success Rate (SR) and Success weighted by Path Length (SPL) are reported as proportions. GDist denotes the remaining Goal-to-Distance in metres.
    \end{threeparttable}
\end{table*}

\begin{figure*}
    \centering
    \includegraphics[width=0.94\textwidth]{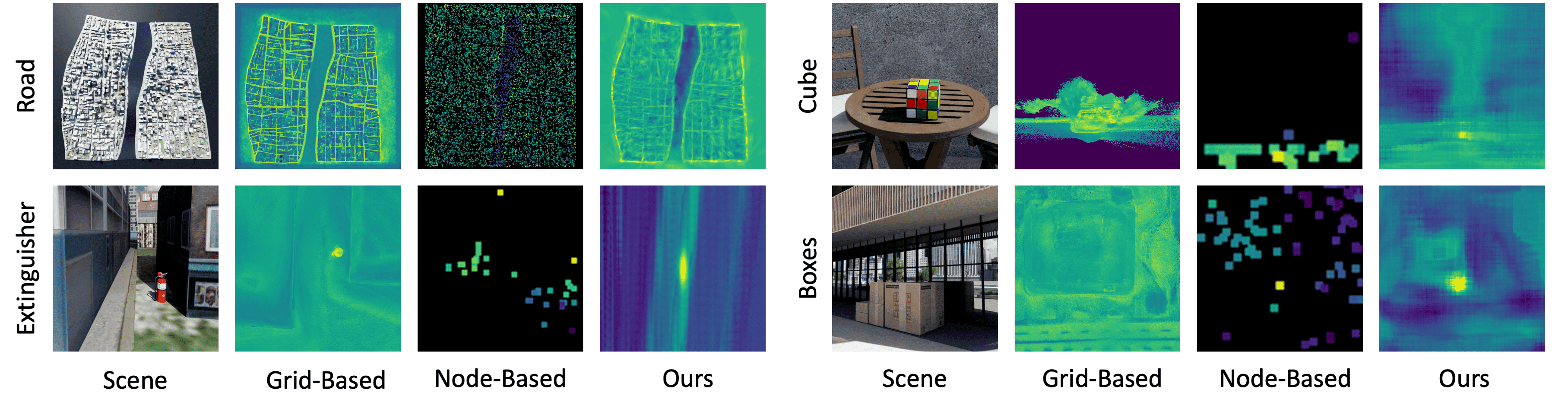}
    \caption{Visualization of each language map representation in the near-goal region of NVIDIA's City Tower Demo 3D Models Pack scene using the viridis colormap. The Node-based method utilizes a dense setting, while the Grid-based method employs a dense setting for Road and Cube scenes and an extremely dense setting (5cm grid size) for Extinguisher and Boxes scenes.}
    \label{fig:vis_result}
\end{figure*}

\subsubsection{Comparison of Language Map Representation Methods for Object Goal Tasks}
As shown in Table~\ref{tab:comparison_methods1}, when using the same amount of memory for map representation, our method outperforms existing approaches in terms of success rate, success weighted by path length, and goal-to-distance metrics. As summarized in Table~\ref{tab:comparison_methods2}, the grid-based method requires approximately 1,000 times more memory than our method to achieve comparable performance by setting the grid size to 40cm. Even with this increased memory usage, the grid-based approach captures large objects but fails to detect smaller ones. In contrast, the node-based method needs about 70 times more memory than our method to reach a similar success rate, yet its performance in the goal-to-distance metric still falls short of that achieved by our implicit method. 
This performance stems from our two-stage pipeline. The initial coarse stage achieves a success rate of 42-67\%, after which the fine-grained optimization stage consistently reduces the goal distance by about 50\%. For easy cases, this optimization always moves closer to the target, and for hard cases, it succeeds in 90\% of cases. Furthermore, while the on-demand embedding generation and fine-path optimization introduce a computational cost, our method completes a full navigation query in under one second, with inference times measured on an NVIDIA RTX 4090 GPU, as shown in Table~\ref{tab:comparison_methods1} and~\ref{tab:comparison_methods2}.

In Fig.~\ref{fig:vis_result}, we present the visualization results of four different methods across distinct scenes: Road, Cube, Extinguisher, and Boxes. For the Road and Cube scenes, the grid-based method employs a dense setting with a grid size of 40cm. However, for a more precise analysis, we adopt a 5cm grid size for the Extinguisher and Boxes scenes, in line with the room-scale settings described in \cite{huang2023visual}. Due to memory constraints in city-scale environments, the 5cm grid size is applied only within specific spatial regions during these experiments.

In the Road scene, all methods effectively represent the spatial layout of the road, as its continuous structure is straightforward to capture. In the Cube scene, however, the grid-based method fails as its 40cm grid resolution is too coarse to capture the smaller cube, causing its representation to be mixed with surrounding objects. In the Extinguisher scene, the node-based method fails because it does not directly observe the goal, whereas our method correctly identifies the target by leveraging the implicit network to infer its presence from the surrounding context. Finally, in the Boxes scene, the grid-based method is hindered by z-axis projection artifacts, while the node-based method detects the boxes but fails to plan a precise path. Our method addresses these issues, accurately detecting the objects and enabling fine-grained goal reaching through optimization.

Overall, these results demonstrate that our method consistently maintains high performance across diverse scenarios by efficiently managing memory and effectively leveraging implicit information, thereby outperforming both grid-based and node-based approaches in challenging environments.

\subsubsection{Performance Comparison of Node Selection Methods}
Selecting appropriate nodes is essential in large-scale graphs to reduce computational complexity while preserving structural properties \cite{zhang2023large, leskovec2006sampling}. In Table~\ref{tab:graph_sampling_comparison}, we compare our method against traditional sampling strategies: Random Node (RN) and Random Degree Node (RDN) \cite{leskovec2006sampling}, which select nodes uniformly at random and based on degree, respectively; Random Walk (RW) \cite{li2015random}, which randomly traverses neighboring nodes; and Forest Fire (FF) \cite{lee2006statistical}, which simulates a spreading process to capture connectivity patterns. In addition, this experiment serves as an ablation study to validate the impact of each component in our proposed scoring mechanism. Unlike conventional methods that rely solely on edge information, our approach leverages implicit information from each node in large graphs, specifically using a neural network to derive an Uncertainty Score ($S_u$) and language embeddings to guide the sampling process. The comparison between rows with and without the $S_u$ term reveals that explicitly modeling uncertainty via our Bayesian framework provides a notable boost in performance, particularly in reducing the final goal to distance (GDist). By integrating the View Coverage Score ($S_{vc}$), Uncertainty Score ($S_u$), and Semantic Sensitivity Score ($S_{ss}$), our method improves SR and SPL, demonstrating the effectiveness of incorporating learned features at the node level to enhance graph sampling beyond classical strategies.

\begin{table}[t!]
    \centering
    \begin{threeparttable}
        \caption{Comparison of node selection methods} 
        \label{tab:graph_sampling_comparison}
        \setlength{\tabcolsep}{8pt}
        \begin{tabular}{lcccccc}
            \toprule
            \textbf{Method} & $\mathbf{S_{vc}}$ & $\mathbf{S_u}$ & $\mathbf{S_{ss}}$ & \textbf{SR} & \textbf{SPL} & \textbf{GDist} \\ 
            \midrule
            RN \cite{leskovec2006sampling}     & --         & --         & --         & 0.48 & 0.44 & 4.34 \\ 
            RDN \cite{leskovec2006sampling}    & --         & --         & --         & 0.45 & 0.40 & 5.31 \\ 
            \midrule
            RW \cite{li2015random}              & --         & --         & --         & 0.48 & 0.43 & 4.13 \\ 
            FF \cite{lee2006statistical}        & --         & --         & --         & 0.45 & 0.41 & 4.79 \\ 
            \midrule
            Ours                               & \checkmark & $\times$   & $\times$   & 0.48 & 0.43 & 4.28 \\ 
            Ours                               & \checkmark & \checkmark & $\times$   & \textbf{0.51} & \textbf{0.47} & 4.36 \\ 
            Ours                               & \checkmark & $\times$   & \checkmark & 0.48 & 0.44 & 4.66 \\ 
            Ours                               & \checkmark & \checkmark & \checkmark & \textbf{0.51} & \textbf{0.47} & \textbf{4.05} \\ 
            \bottomrule
        \end{tabular}
        $\mathbf{S_{vc}}$, $\mathbf{S_u}$, and $\mathbf{S_{ss}}$ refer to the View Coverage Score, Uncertainty Score, and Semantic Sensitivity Score, respectively, as detailed in Section~\ref{subsec:B}. 
    \end{threeparttable}
\end{table}

\subsection{Real-World Experiments and Analysis}
\label{subsec_ex:C}
\begin{figure*}
    \centering
    \includegraphics[width=0.80\textwidth]{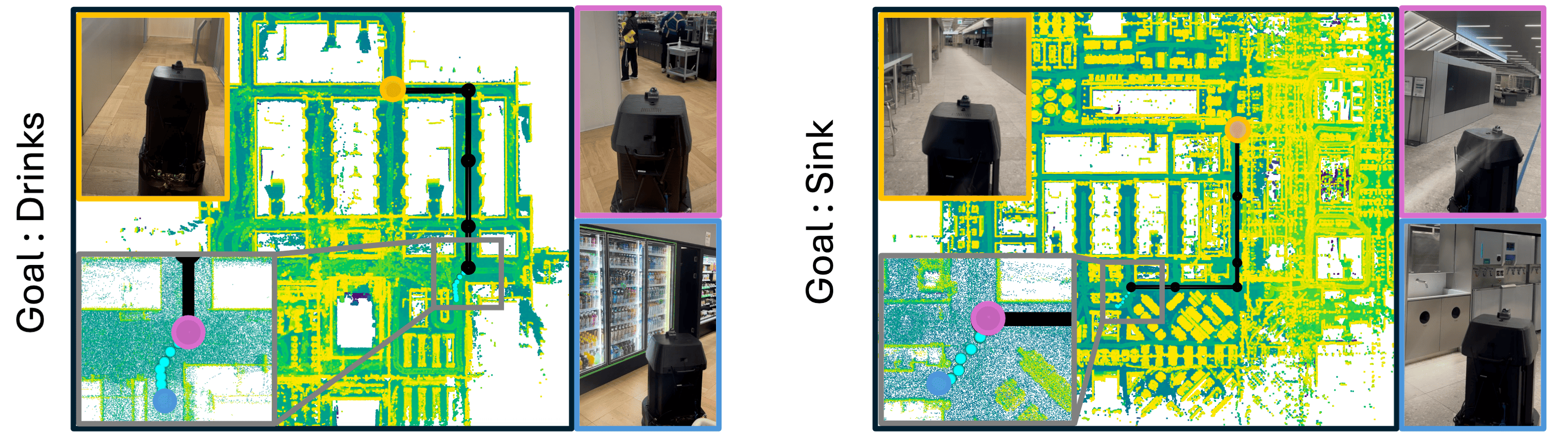}

    \caption{Visualization of real-world experiments. 
        \startsymbol~: Start pose, representing the initial position from which navigation is initiated. 
        \coarsesymbol~: Coarse goal pose, selected based on its language embedding similarity to the target object. 
        \optsymbol~: Optimized goal pose, which is obtained through optimization of the implicit language map, ensuring that the target object is accurately brought into view.}
    
    \label{fig:realworld}
\end{figure*}

We conduct real-world experiments using the M2 robot \cite{naverlabsM2} for zero-shot object goal navigation. Our real-world experiments evaluate our method's precision using a sparse graph and its ability to navigate towards goals not directly observed by any node. To do this, we set approximately 20–30 nodes sparsely on each floor. During navigation, the robot first moves toward a target node based on the given natural language command and then refines its position through optimization. Specifically, when navigating to “drinks", no node in the map directly observes the drinks; however, as shown in Fig.~\ref{fig:realworld}, a node near the drinks is selected, and through subsequent optimization, the robot adjusts its position to view the drinks. The implicit map learns from a large number of pose-image pairs and continuously represents coordinates in the language feature space, allowing nodes that have not directly observed the goal to possess language vectors similar to those of the goal. Furthermore, during the process of moving to the visible location of the goal, the method utilizes gradients in the continuous language space, enabling path planning to the actual visible location of the goal. These results demonstrate that our method can effectively represent space, allowing the robot to precisely locate target objects without relying on memory-intensive explicit language maps.
To quantitatively substantiate these findings, we report the key performance metrics from our real-world trial in Table~\ref{tab:real_world_quant}, where LAMP significantly outperforms the explicit baseline in goal-reaching accuracy.

\begin{table}[t]
    \centering
    \begin{threeparttable}
        \caption{Real-World Navigation Performance}
        \label{tab:real_world_quant}
        \setlength{\tabcolsep}{6pt} 
        \begin{tabular}{lccc}
            \toprule
            \textbf{Method} & \textbf{Success (\%)} & \textbf{GDist (m)} & \textbf{Time (s)} \\
            \midrule
            Explicit Baseline & 90.0 & 5.19 & 1.82 \\
            Ours (LAMP)       & 90.0 & 1.89 & 3.84 \\
            \bottomrule
        \end{tabular}
        Real-world trial results. The explicit baseline uses pre-computed CLIP features for each node. The time reflects the total end-to-end duration for planning each query.
    \end{threeparttable}
\end{table}

However, this approach has some limitations. Its effectiveness depends on the VLM’s accuracy. For instance, if other objects share similar visual features with the target, the model may mistakenly select wrong nodes. Moreover, when the target’s appearance is weak or ambiguous, it might struggle to identify the correct nodes. Therefore, enhancing the VLM's performance and selecting objects with distinct and clear features are crucial for improving navigation accuracy.

\section{CONCLUSIONS}
We introduce LAMP, an implicit language map for large-scale, zero-shot navigation using only RGB images. Beyond a simple implicit representation, LAMP is the first to leverage a language-driven continuous field for fine-grained path generation. Our approach combines a continuous neural network with a sparse topological graph, achieving efficient memory usage by dynamically generating language embeddings for path planning rather than storing them explicitly. Additionally, we enhance pose refinement by sampling poses near the coarse estimate and adding a distance penalty to avoid local minima. We also model embedding uncertainty in a Bayesian framework and use learned node-level features for efficient graph sampling. Simulation experiments demonstrate that LAMP surpasses traditional grid- and node-based approaches in diverse scenarios, offering improved or comparable performance with substantially reduced memory requirements. Real-world experiments show that continuously encoding language features enables LAMP to detect and navigate toward objects not explicitly observed during mapping.

{
\footnotesize
\bibliographystyle{IEEEtran}
\bibliography{main}
}

\end{document}